\documentclass{article}
\usepackage[preprint]{spconf}
\usepackage{amsmath,graphicx}

\usepackage{svg}
\usepackage{amsmath}
\usepackage{float}
\usepackage{graphics}
\DeclareMathOperator*{\argmax}{arg\,max}

\usepackage[multiple]{footmisc}
\usepackage{multirow}
\newcommand{\mc}{\multicolumn}

\copyrightnotice{\copyright\ IEEE 2020}
\toappear{To appear in {\it Proc.\ SLT 2021, January 19-22, 2021, Shenzhen, China.}}

\title{VirAAL: Virtual Adversarial Active Learning for NLU}
\name{Gregory Senay\sthanks{\ Authors made equivalent contributions to this work.}, Badr Youbi Idrissi\footnotemark[1]\sthanks{\ Work done during internship at xBrain.}, Marine Haziza} \address{ Augustus Intelligence/xBrain, Menlo Park, CA, US.}

\begin{document}

\maketitle
\begin{abstract}
This paper presents VirAAL, an Active Learning framework based on Adversarial Training.
VirAAL aims to reduce the effort of annotation in Natural Language Understanding (NLU).
VirAAL is based on Virtual Adversarial Training (VAT), a semi-supervised approach that regularizes the model through Local Distributional Smoothness. 
With that, adversarial perturbations are added to the inputs making the posterior distribution more consistent. 
Therefore, entropy-based Active Learning becomes robust by querying more informative samples without requiring additional components. 
The first set of experiments studies the impact of an adapted VAT for joint-NLU tasks within low labeled data regimes. 
The second set shows the effect of VirAAL in an Active Learning (AL) process. 
Results demonstrate that VAT is robust even on multi-task training, where the adversarial noise is computed from multiple loss functions.
Substantial improvements are observed with entropy-based AL with VirAAL for querying data to annotate.
VirAAL is an inexpensive method in terms of AL computation with a positive impact on data sampling. 
Furthermore, VirAAL decreases annotations in AL up to 80\% and shows improvements over existing data augmentation methods.
The code is publicly available.
\end{abstract}

\section{Introduction}
Data annotation is time-consuming, expensive, and often requires experts or at least a good understanding of the data to reach a qualitative annotation. 
Scalable, fast, and cheap methods exist, such as crowdsourcing solutions, like Amazon Mechanical Turk (AMT).
However, using AMT or third parties is impossible when data privacy matters and must require internal annotation.
This is especially true in specific domains such as banking, insurance, or medical.
Nevertheless, there are more and more available services with user interfaces, like spoken dialog systems.
These systems must be trained on large datasets in order to achieve acceptable interactions with users.
They rely on Speech or Natural Language Understanding (SLU, NLU) \cite{demori2008spoken,DBLP:books/daglib/p/Bellegarda14}.
In addition to user intents, annotations can include slot and optional domain information, making multi-task annotation complex, time-consuming, and repetitive.

Dialog systems can collect massive amounts of user data, but this data can rarely be used directly and very often is impossible to annotate.
One way to take advantage of these amounts of data is to use semi-supervised approaches \cite{zhu2005semi, chapelle2009semi, zhu2009introduction}.
These methods have yielded consistently reliable results in text classification \cite{Miyato2016VirtualAT, revisitinglstm, MixMatch} for datasets like IMDB, Rotten Tomatoes, DBpedia, or RC1.
However, these studies usually focus on large labeled datasets.

On the other hand, Active Learning aims to minimize the number of manual annotations required to reach an acceptable performance level.
Active learners query unlabeled samples to be annotated \cite{Settles2009ActiveLL, Ferreira_Bandit, Garcia_AL}.
These queries are designed to extract informative or diverse samples.
Some popular methods are uncertainty-based queries.
Furthermore, many studies \cite{Ducoffe2018AdversarialAL, Gissin2018DiscriminativeAL} show that random query selection is a robust and consistent baseline.
Usually, efficient methods are too complex \cite{Ducoffe2018AdversarialAL,Gissin2018DiscriminativeAL,sener2018active,BALD} to be easily integrated into the annotation process, making it time-consuming and expensive to be deployed effectively.
Additionally, these methods that sample data to be annotated do not contribute directly to the training.

Among all these methods, Virtual Adversarial Training (VAT) \cite{Miyato2016VirtualAT, Miyato_Smoothing}, a semi-supervised approach, proves its robustness in a large variety of contexts like computer vision, text classification, or speech \cite{VAT, Miyato2016VirtualAT, revisitinglstm, Shinohara2016}.
VAT regularizes the training by adding adversarial examples and avoids overfitting.
It could be interesting to evaluate its impact on NLU in a low data regime.
Nevertheless, recent advances in NLU \cite{NLU_joint_att_Bing, nlu-slotgated, hakkani-tr2016multi-domain} rely on joint optimization of intent detection and slot filling, while VAT has been designed for single tasks and mainly evaluated on large datasets.
Besides, if VAT can correctly regularize the model, even in a low data regime, it will then be possible to obtain better confidence in the model posterior distribution, which would therefore make it possible to query unlabeled samples better using the improved model confidence. 
This would occur without using additional complex Active Learning methods and would increase the NLU model generalization with fewer data while reducing the effort of human annotation by querying more informative data.

This paper tries to answer these questions by evaluating the effectiveness of VAT in low data regimes in Spoken Language Understanding and its impact on the posterior distribution in an Active Learning paradigm.
This framework is called VirAAL for Virtual Adversarial Active Learning.
VirAAL combines VAT with uncertainty-based Active Learning in an attempt to increase sample efficiency even more. 
Section \ref{viraal} presents VirAAL and its components and methods for NLU: a joint-NLU model, an adapted Virtual Adversarial Training and the Active Learning procedure.
Section \ref{expe} shows the protocol, experiments and results for low data regimes, Active Learning and a comparison with existing data augmentation methods.
The paper ends with a conclusion and some perspectives.
\section{VirAAL}
\label{viraal}
\begin{figure*}[ht]
  \begin{minipage}[b]{1.0\linewidth}
  \centering
  \includegraphics[width=0.80\linewidth]{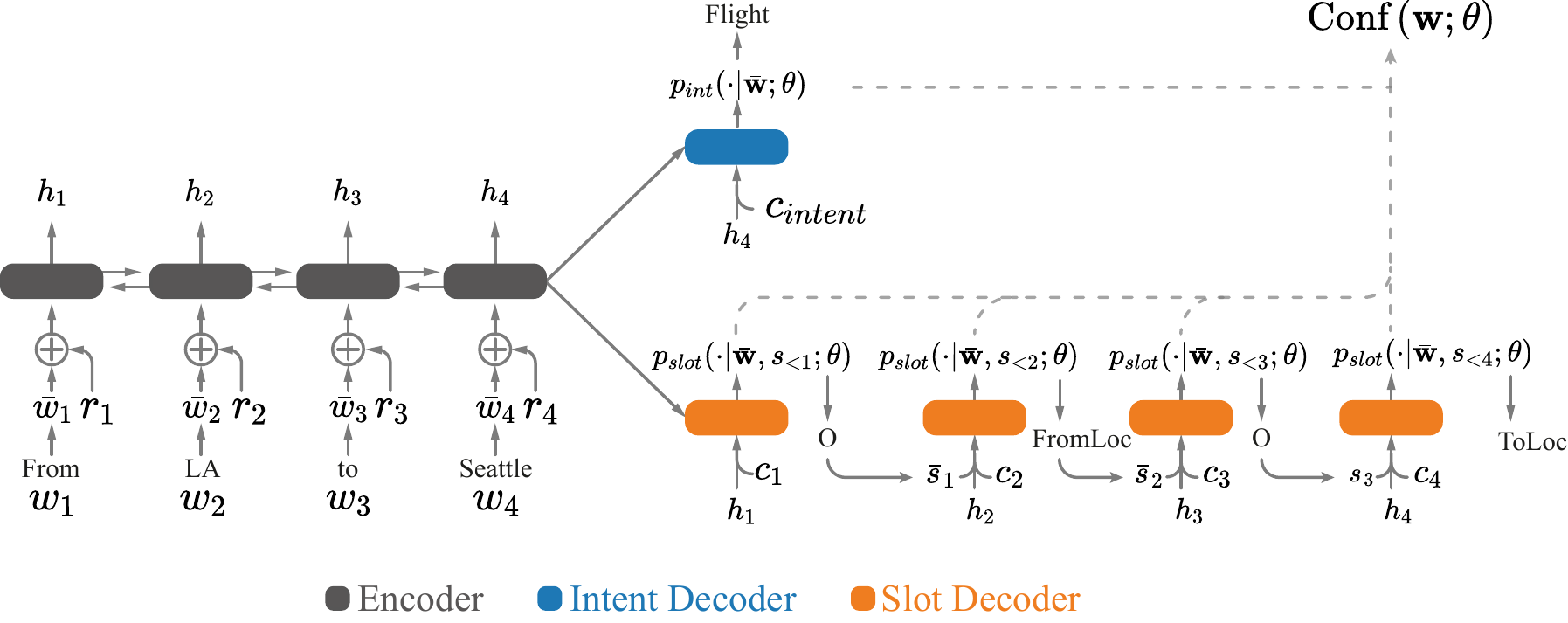}
  \caption{VirAAL: an Attention-based Recurrent Neural Network model for joint-NLU (intent detection and slot filling) \cite{NLU_joint_att_Bing} where adversarial noise $r_t$ is added to the embeddings $\bar{w}_t$. Active Learning querying functions are based on entropy.} 
  \label{fig:joint_model}
  \end{minipage}
\end{figure*}

Virtual Adversarial Active Learning (VirAAL) relies on three components: a joint-NLU neural network model for joint intent detection and slot filling, a semi-supervised training method based on Virtual Adversarial Training adapted for joint training and entropy-based querying functions for Active Learning.
The next sections describe the model, the methods and our motivations.
Figure \ref{fig:joint_model} illustrates VirAAL with the adversarial noise on inputs and entropy-based Active Learning criteria.

\subsection{Joint-NLU}
In recent years, joint training has been shown to accomplish state-of-the-art results in NLU \cite{nlu-slotgated, hakkani-tr2016multi-domain, yang_e2e, chen2019bert}. 
The model architecture used in VirAAL is an Attention-based Recurrent Neural Network for joint intent detection and slot filling \cite{NLU_joint_att_Bing} with aligned inputs.
This model is relatively light and fast to train for iterative human-in-the-loop training while providing good results in NLU. 
Let's denote $\mathbf{w}=\{w_t|t=1,...,T\}$ an input sequence of words, where $T$ is the sequence length and $\mathbf{s}=\{s_t|t=1,...,T \}$ the target sequence of slots.
$\bar{\mathbf{w}}$ and $\bar{\mathbf{s}}$ are the corresponding embeddings of $\mathbf{w}$ and $\mathbf{s}$.
$\mathrm{i} \in \mathcal{I}$  is the target intent.
The superscript $x^{(k)}$ denotes the $k$-th element of $K = K_l+K_u$ where $K_l$ is the labeled set and $K_u$ the unlabeled set.

The model has a bidirectional-LSTM \cite{HochreiterLSTM} sentence encoder taking word embeddings $\bar{\mathbf{w}}$ as input, and generates $\mathbf{h}=\{h_t|t=1,...,T\}$ a hidden representation.
The intent attention decoder takes a context vector $c_{intent}$ that is a weighted sum of the encoder hidden states $\mathbf{h}$ with the last real word hidden state $h_T$ as the attention key.
The slot filling decoder is a unidirectional LSTM that takes as input the aligned encoder output $\mathbf{h}$, the context vector $c_t$, and the last predicted embedding $\bar{s}_{t-1}$ slot.
$c_t$ is an attention vector where the key is $\bar{s}_{t-1}$. 
The decoders have a softmax layer to output the estimated probability distribution of the targeted intent $p_{int}(\mathrm{i}|\bar{\mathbf{w}}; \theta)$ and slots  $p_{slot}(s_t|\bar{\mathbf{w}}, s_{< t}; \theta)$. 
The intent and slot filling losses are defined by:
\begin{equation}
\mathcal{L}_{int}(\theta) = - \frac{1}{K_l}\sum_{k=1}^{K_l} \log(p_{int}(\mathrm{i}^{(k)}|\bar{\mathbf{w}}; \theta))
\\ 
\end{equation}
\begin{equation}
\mathcal{L}_{slot}(\theta) = - \frac{1}{K_lT}\sum_{k=1}^{K_l} \sum_{t=1}^{T}\log(p_{slot}(s_t^{(k)}|\bar{\mathbf{w}}, s_{< t}^{(k)}; \theta))
\end{equation} 
Finally, the joint training aims to minimize the summed loss functions:
$\mathcal{L}(\theta) = \mathcal{L}_{int}(\theta) + \mathcal{L}_{slot}(\theta)$.

\subsection{Virtual Adversarial Training}
Neural networks have been shown to exhibit a very peculiar property: they are very sensitive to adversarial examples \cite{Goodfellow2014ExplainingAH}.
These samples are crafted by adding small perturbations optimized to mislead the model into predicting a wrong label.
In the case of computer vision models, these perturbations are almost imperceptible to the human eye.
Various methods were suggested to counter this effect \cite{Carlini2019OnEA}, including adversarial training \cite{madry2018towards}.
This method consists of augmenting the dataset with adversarial examples to make the model more robust.
Virtual Adversarial Training \cite{VAT} extends this notion by replacing the ground truth with the best approximation of it: the model output probability distributions.
Virtual Adversarial Training (VAT) is a regularization method that encourages Local Distributional Smoothness (LDS) \cite{Miyato_Smoothing}.
Since VAT does not require all samples to be labeled, it can be used in a semi-supervised context. 
However, due to the discrete aspect of words in NLP, these small perturbations can be added to the word embedding vectors $\bar{\mathbf{w}}$. 
Virtual Adversarial Training for NLP \cite{Miyato2016VirtualAT} aims to improve the generalization rather than defend against adversarial examples.

Formally, this method consists of minimizing the Kullback-Leibler (KL) divergence between the posterior distribution given $x$ and the adversarial distribution given $x+r_{vadv}$:
\begin{equation*}
    \mathrm{LDS}(x) = \mathrm{D}(p, x, r_{vadv}) = \mathrm{KL} \left[ {p(\cdot|x,\hat{\theta})\bigg|\bigg|p(\cdot|x+r_{vadv},\theta)} \right]
\end{equation*}
where $\hat{\theta}$ is a constant copy of the parameters $\theta$ of the model. $\mathrm{D}(p,x,r)$ is the KL divergence of probability distribution p between point $x$ and $x+r$.
$\mathrm{LDS}$ measures to which extent the posterior distribution is locally smooth around $x$. $r_{vadv}$ satisfies:
\begin{equation*}
r_{vadv} = \argmax_{r,\Vert{r}\Vert_2 < \epsilon}{\mathrm{D}(p,x, r)}
\end{equation*}
Since calculating this exact perturbation is too computationally heavy, \cite{VAT} approximates it with $r_{vadv} = -\epsilon g/\Vert{g}\Vert_2$ where:
\begin{equation*}
g=\nabla_{r}\mathrm{D}(p, x, r)
\end{equation*}
Back-propagation is not done through $r_{vadv}$ when training with VAT.

Here, VAT is adapted for the joint-NLU training.
A single perturbation is computed for both intent detection and slot filling.
Since the two heads share the same encoder parameters, using two separate perturbations could lead to an inconsistent update.
The combined perturbation $r_{joint}$ is computed by taking the gradient of $\frac{1}{2}\left(\mathrm{D}_{int}(x,r)+\mathrm{D}_{slot}(x,r)\right)$.
With 

\begin{equation*}
 \mathrm{D}_{int}(\bar{\mathbf{w}},r) = \frac{1}{K}\sum_{k=1}^{K}  \mathrm{D}(p_{int},\bar{\mathbf{w}}^{(k)},r)
\end{equation*}
Since slot filling is a sequence-to-sequence task, $\mathrm{D}_{slot}$ is calculated on every time step $t$ and then averaged:
\begin{equation*}
\mathrm{D}_{slot}(\bar{\mathbf{w}},r) = \frac{1}{N}\sum_{k=1}^{K} \sum_{t=1}^{T}\mathrm{D}(p_{slot},\bar{\mathbf{w}},r) 
\end{equation*}
And finally, the joint VAT loss is
$$
    \mathcal{L}_{vat}(\theta) = LDS_{joint}(\bar{\mathbf{w}}) = \frac{1}{2}\left(\mathrm{D}_{int}+\mathrm{D}_{slot}\right)(\bar{\mathbf{w}},r_{joint})
$$
with $r_{joint} = \frac{1}{2}(\nabla_r D_{int}(\bar{\mathbf{w}}, r)+\nabla_r D_{slot}(\bar{\mathbf{w}}, r)) = \frac{1}{2}(r_{int}+r_{slot})$.
When the perturbation directions cancel out, the joint VAT loss has no effect.
Indeed if $r_{int}=-r_{slot}$, then $r_{joint}=0$, and thus $\mathcal{L}_{vat}(\theta)=0$.
On the other hand if $\mathcal{L}_{vat}^{int}(\theta)$ and $\mathcal{L}_{vat}^{slot}(\theta)$ are separately minimized on the same step, the encoder gradients could become inconsistent with the two decoder head gradients. 
In the limits of the first-order approximation, minimizing the joint vat loss decreases both KL divergences or has no effect.
However, minimizing the two losses separately is slower (requires additional forward and backward passes) and does not necessarily decrease both KL divergences.
The final joint loss is defined as:
\begin{equation}
 \mathcal{L}(\theta) = \mathcal{L}_{int}(\theta) +\mathcal{L}_{slot}(\theta)+ \mathcal{L}_{vat}(\theta)
\end{equation}
\subsection{Active Learning}
\label{active_learning}
Prominent studies show that deep neural networks are poorly calibrated \cite{pmlr-v70-guo17a, NIPS2017_7219}.
Calibration estimation reflects how much the model is representative of the true correctness likelihood.
Models often output high probabilities on samples that are far from the training distribution.
Through LDS, VAT could improve uncertainty-based Active Learning as evidenced by its ability to output better-calibrated probabilities \cite{revisitinglstm}, and thus, more informative samples could be queried as a result.

In VirAAL, Active Learning querying functions are based on the entropy of the output distributions, rather than using only the predicted class probability.
Intuitively, this allows to estimate how much information the distribution contains in the sample $x$ and hence, estimate how confident it is in its predictions.
It also results in a less-complex method by estimating the samples individually.
Contrary to Batch Active Learning methods that provide better sample diversity\cite{Kirsch2019BatchBALDEA}, VirAAL evaluates each sample individually to keep a linear complexity.
In the case of live human interactions for labeling, minimizing the waiting time is crucial.
As the system suggests new samples to be labeled, it also ensures efficient interactions with the human labeling the data.
The entropy of a probability distribution is defined as follows:
\begin{equation*}
H(p(\cdot)) = -\sum_{x\in\mathcal{X}} p(x)\log(p(x) )
\end{equation*}
The intent prediction confidence given a sample $\mathbf{w}$ is
\begin{equation}
\label{eq:conf_int}
\text{Conf}_{int}(\mathbf{w};\theta) = -H(p_{int}(\cdot|\bar{\mathbf{w}};\theta))
\end{equation}
Because slot filling tasks produce a posterior distribution for each word, slot filling confidence of an utterance $\mathbf{w}$ is based on the mean-entropy:
\begin{equation}
\label{eq:conf_slot}
\text{Conf}_{slot}(\mathbf{w};\theta) = \frac{1}{T}\sum_{t=1}^T -H(p_{slot}(\cdot|\bar{\mathbf{w}}, s_{< t}; \theta))
\end{equation}
This confidence estimation is similar to the Mean Negative Log Probability (MNLP) \cite{shen2018deep, siddhant-lipton-2018-deep} used in Named Entity Recognition.
Finally, the joint (intent detection and slot filling) Active Learning confidence score is defined by the joint confidence, using the normalized confidence measures by the 99th percentile of each score.
The $S$ selected samples to be labeled are those with the lowest confidence scores.
\begin{equation}
\label{eq:conf_joint}
\text{Conf}(\mathbf{w};\theta)=\frac{\text{Conf}_{int}(\mathbf{w};\theta)}{P^I_{99}}+\frac{\text{Conf}_{slot}(\mathbf{w};\theta)}{P^S_{99}}
\end{equation}

\section{Experiments}
\label{expe}
Experiments are conducted in low data regimes: only a small set of labeled data is available.
Three sets of experiments are presented.
The first set of experiments is the impact of the VAT on intent detection only, slot filling only and then joint training.
Then, the Active Learning experiments aim to show the impact of VAT on the Active Learning framework VirAAL, where the annotator is supposed to label two sets of data.
At first, the initial set is built with a random data sample.
This set is then labeled, followed by a first training with and without VAT.
Experiments present the final training performed with an additional set selected with two criteria: Random or Entropy.
This second set size is equal to the first, doubling the labeled training data.
Notably, this paper does not focus on the initial sampling selection.
The third experiment is conducted to compare VirAAL methods to existing data augmentation methods.

\subsection{Datasets}
\label{datasets}
Experiments are performed on two Spoken Language Understanding datasets: ATIS and SNIPS.
ATIS (Airline Travel Information Systems) dataset \cite{atis} contains recordings for reserving flights and is widely used in SLU research.
This work follows the data split used in many papers \cite{NLU_joint_att_Bing, Mesnil:2015:URN:2817174.2817185, Tur_atis, Xu_atis,nlu-slotgated}. 
The training set contains 4,978 utterances and is composed of 127 slots and 18 different intents.
A development set of 500 utterances is extracted from it, and the test set is composed of 893 utterances.
SNIPS dataset \cite{SNIPS} is a voice assistant dataset and contains data annotated in 7 intents (Music, Book, Weather...) and 73 slots.
SNIPS is composed of a training set of 13,084 utterances and development and test sets with 700 utterances within each.

\subsection{Protocol}
\label{protocol}

In the first experiment, ATIS is evaluated at $5\%$ and all multiples of $10\%$ of labeled data ($5\%$ of ATIS labeled data corresponds to 230 labeled utterances). 
SNIPS is evaluated at every $1\%$ between $0\%$ and $10\%$, and all multiples of $10\%$ of labeled data.
In SNIPS, $1\%$ of labeled data corresponds to 131 labeled utterances. 
This experiment compares trainings using 6 different losses: int; slot; joint; (vat, int); (vat, slot); (vat, joint).
In the Active Learning experiments, the different criteria are evaluated for different set sizes: $\{10,20,40,50,60\}\%$ for ATIS and $\{2,4,6,8,10,20,40\}\%$ for SNIPS.
In Figure \ref{fig:rerank}, $X\%$ corresponds to an initial training set of $\frac{X}{2}\%$ randomly annotated data.
Then, an additional $\frac{X}{2}\%$ are selected by one of the methods.
The randomly sampled set contains at least 1 of each intent to be similar to real-world scenarios, regardless of the slot type coverage: some slot types can be missing.
Additionally, in Active Learning, the initial random set is the same for all the different methods.

Table \ref{tab:hyperparamters} presents the hyper-parameters used in all experiments.
It should be noted that experiments use different batch sizes between non-VAT and VAT trainings for the ATIS experiments.
The larger batch size in VAT aims to include both labeled and unlabeled data in each mini-batch to optimize all losses simultaneously.
Following previous works in joint-NLU \cite{NLU_joint_att_Bing}, a batch size of 16 is used on non-VAT trainings.
All SNIPS and VAT trainings use a 64 batch size.
Non-VAT trainings have been tested on smaller batch sizes showing no improvement or even worse performances.
Evaluation metrics are accuracy for intent detection and token-level micro-average F1 score for slot filling.
Furthermore, all experiments present means and standard deviations on 8 runs.
A validation set proportional to the data regimes has been used to ensure no overfitting.
Additionally, for reproducibility, the code is available\footnote{https://github.com/xbraininc/viraal/} and experiments use FastText embeddings.

\begin{table}[th]
\centering  
\begin{tabular}{lr|lr}
\hline
    Parameter & value           & Parameter & value\\
\hline
    Embedding size & 300		& Epochs & 100/60\\
    LSTM hidden size & 128		& Batch sizes & 16/64\\
    LSTM layer & 1 				& Optimizer & Adam\\
    Slot emb. size & 128 	    & Learning rate & 0.001\\
    Classifier Dropout & 0.5    & Emb. Dropout &0.5 (VAT 0)\\
\hline
\end{tabular}
\caption{Training hyper-parameters used in all experiments.}
\label{tab:hyperparamters}
\end{table}

\subsection{Results}
\label{results}

\begin{figure}[htp]
\begin{minipage}[h]{1.0\linewidth}
  \centering
  \includegraphics[width=\linewidth]{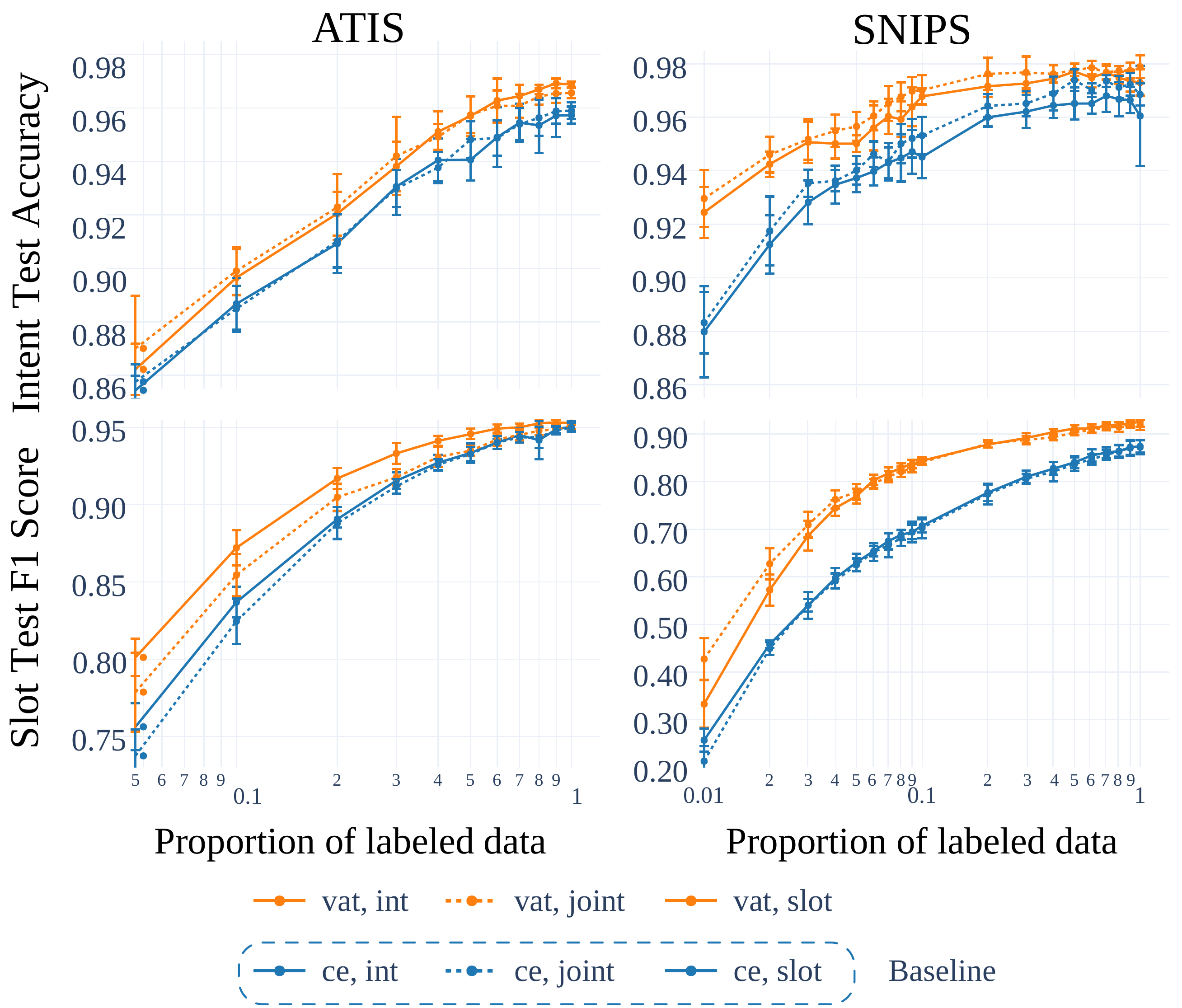}
  \caption{Test scores of NLU trainings. Left column is ATIS, right column is SNIPS. First row shows the intent accuracy and second row the slot filling F1 score. The model is either trained with intent (int) only, slot filling (slot) only or joint training (joint). The training losses are cross-entropy (ce) or virtual adversarial (vat).}
  \label{fig:train}
  \end{minipage}
\end{figure}

Figure \ref{fig:train} shows consistent improvements of VAT trainings (in orange) across the different labeled data regimes against non-VAT training (in blue).
The most significant improvements occur in the low data regimes (less than 30\%) for all training types: intent detection, slot filling and joint.
This confirms the hypothesis that VAT, even used in a low labeled data regime, is able to better regularize the model using the unlabeled data.
This suggests it is due to the smoothing of the decision boundary by VAT that propagates the labeled signals from labeled samples to the neighboring unlabeled samples.
This is even more apparent in low data regimes where the labeled signal is sparse.
This, in itself, is a form of Active Learning as \cite{VAT} pointed out.
Additionally, VAT joint training outperforms the baseline, which shows that regularizing with adversarial noise from two different signals (VAT slot loss and VAT intent loss) is still efficient and suggests it could be used in other multi-task training.
More interestingly, VAT joint training in 60\% and 10\% regimes achieves very similar intent accuracies as 90\% (+0.2\%) and 50\% (-0.4\%) without VAT respectively in ATIS and SNIPS datasets.

\begin{figure}[hb!]
  \centering
  \centerline{\includegraphics[width=0.95\linewidth]{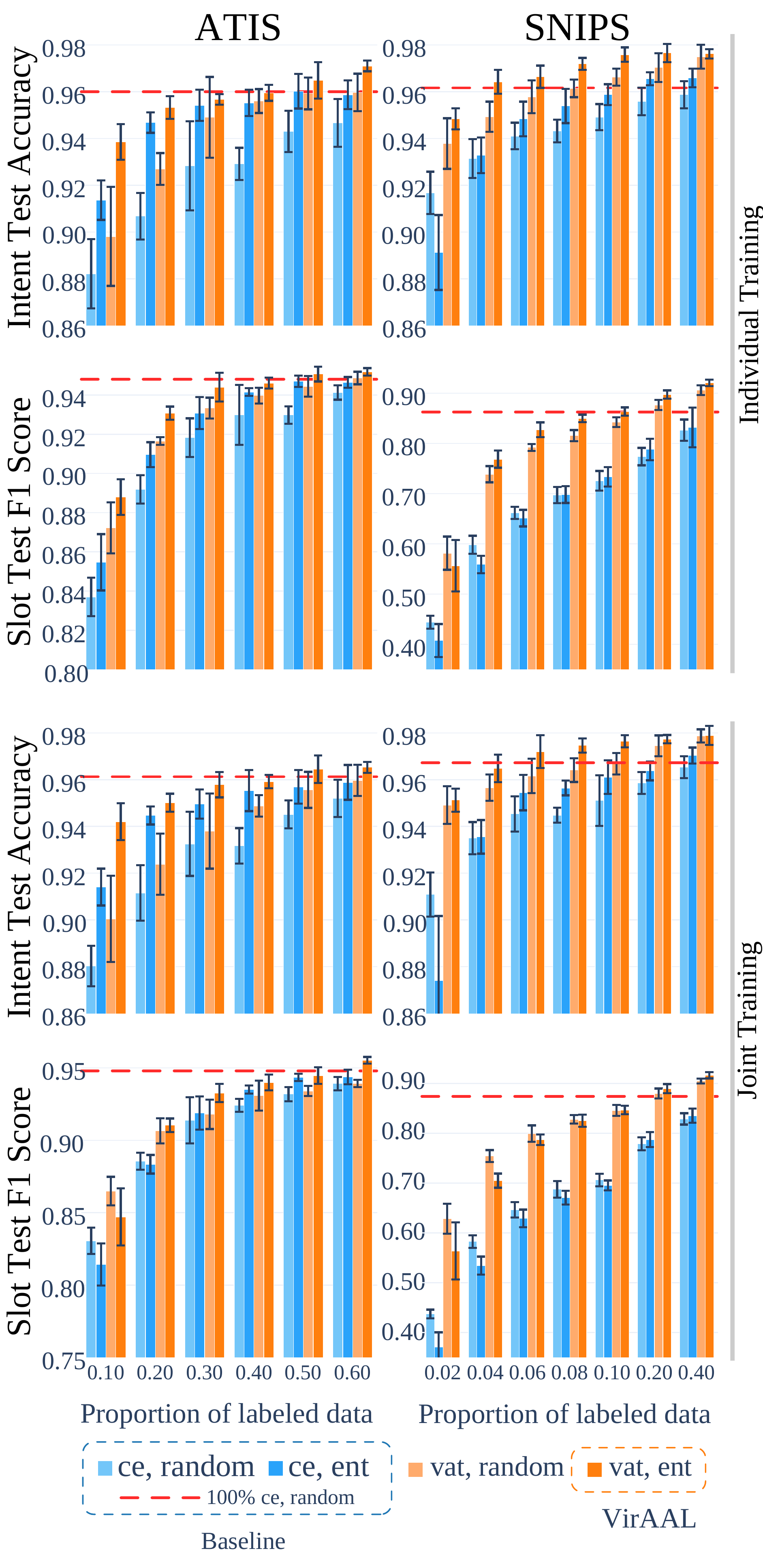}}
  \caption{VirAAL (vat, ent) and (vat, random) results on ATIS (left) and SNIPS (right) datasets obtained on low data regimes for the 3 different trainings: intent, slot filling and joint. Active Learning criteria are: random, $\text{Conf}_{int}$, $\text{Conf}_{slot}$ and \text{Conf} for joint training.}
  \label{fig:rerank}
\end{figure}

Figure \ref{fig:rerank} shows even greater improvements with Active Learning (AL).
The largest gains are obtained in intent detection with or without joint training.
In SNIPS, only 10\% of labeled data with VirAAL (vat, ent) is enough to have higher scores than the 100\% regime without VAT and AL (lines in red).
This can reduce the annotation effort by up to 80\%.
More precisely, intent accuracy reaches 97.89\% and slot F1 88.95 with VirAAL versus 96.86\% and 87.40 F1, in joint training.
A similar improvement is shown in ATIS with a 30\% regime, respectively, with +0.62\% and +1.54 F1 of absolute differences.
On intent training only on SNIPS, 2\% of labels suffices for the intent accuracy to overtake the 100\% regime: 96.43\% versus 96.05\%.
Furthermore, VirAAL for intent detection always outperforms the random baseline (vat, random) or AL without VAT (ce, ent).
VirAAL outperforms random in almost all data regimes and for both datasets contrary to entropy-based AL without VAT, yielding worse scores than random in many regimes.
Nevertheless, VirAAL in joint training shows mitigated results with slight improvements and a similar behavior as joint (ce, ent).
This is probably due to the non-heterogeneous natures of intent and slot scores for AL: slot confidence is an average of the slot entropies.
In that respect, it would be interesting to use AL criterion in a common latent space.

\subsection{Comparison with existing NLU methods in low data regimes}
\begin{table*}[th!]
\centering  
\begin{tabular}{|l|c|c|c|c|c|c|c|c|}
\hline
\textbf{Dataset} &  \mc{4}{|c|}{ATIS} & \mc{4}{|c|}{SNIPS}\\ \hline
\textbf{Split} &\mc{2}{|c|}{Small}  &  \mc{2}{|c|}{Medium} & \mc{2}{|c|}{Small} & \mc{2}{|c|}{Medium} \\ \hline
\textbf{Model}          & Slot    & Int   &  Slot &  Int  & Slot    & Int   &  Slot   & Int    \\ \hline\hline
 joint K-SAN (syntax)
                        &    74.56 &    -     &    88.40 &    -     &    -     & -     & -     & -      \\ \hline
JLUVA + Additive
                        &    74.14 &    83.46 &    89.13 &    90.97 &    -     & -     & -     & -      \\ \hline
JLUVA + Posterior
                        &    74.92 &    83.65 &    89.27 &    90.95 &    -     & -     & -     & -      \\ \hline
SC-GPT-NLU \textit{Paired-Data-Only}
                        &    75.42 &    86.67 &    88.61 &    90.71 &    64.96 &   93.43  &    80.62 &    97.57  \\ \hline
SC-GPT-NLU \textit{Rich-in-Utterance}
                        &    78.45 &\bf{87.46} &    88.23 &    91.94 &    63.46 &   93.43  &    80.54 &\bf{98.14}  \\ \hline\hline
Baseline                &    72.19 &    84.53 &    87.19 &    90.20 &    65.59 &   94.25  &    82.19 &    95.89  \\ \hline
VAT Joint (Our)
                        &    75.61 &   85.13  &    89.33 &    91.21 &    67.15 &   94.70  &    85.25 &    96.41  \\ \hline
VirAAL joint-entropy (Our)
                        &    71.60 &    86.23 &   88.78  &\bf{94.58}&    63.26 &\bf{95.52}&   85.61  &    97.61 \\ \hline 
VirAAL individual-entropy (Our)
                        &\bf{78.47}&   86.55  &\bf{90.33}&    94.09 &\bf{69.48}&    95.34 &\bf{87.85}&    97.27 \\ \hline 
\end{tabular}
\caption{VirAAL and VAT joint methods compare to data augmentation methods on ATIS and SNIPS, Small and Medium test sets. Int is the intent accuracy, Slot is the slot F1. VirAAL joint-entropy corresponds to Eq. (\ref{eq:conf_joint}), individual-entropy to (\ref{eq:conf_int}) or (\ref{eq:conf_slot}).}
\label{tab:data_augmentation}
\end{table*}
 
To evaluate and compare our methods in a low data regime, additional experiments are conducted with a specific protocol.
Following \cite{chen2016syntax}, VirAAL methods are evaluated on ATIS and SNIPS \textit{Small} and \textit{Medium}.
It corresponds respectively to (1/40) and (1/10) of the full datasets.
More precisely, ATIS Small and Medium represent 129 and 515 utterances, while SNIPS Small and Medium 327 and 1308 utterances.
To be as close as possible to the existing data augmentation setups \cite{chen2016syntax,JLUVA,peng2020data} and for a fair comparison, VirAAL pre-trained embeddings are initialized with 300-dimension Glove vectors \cite{pennington-etal-2014-glove}.
Moreover, batch sizes $\{4, 8, 16, 32, 64\}$ and the number of epochs are tuned on the full validation sets.
This is in contrast to the main experiments of section \ref{protocol}, where only a proportional validation is used to simulate a realistic training scenario, resulting in a high variance.
Apart from that, all the parameters are identical as Table \ref{tab:hyperparamters}.

VirAAL methods are compared in Table \ref{tab:data_augmentation} with the with the following baselines:
K-SAN syntax \cite{chen2016syntax} does not use data augmentation; instead, the augmentation is done using syntactic information.
JLUVA \cite{JLUVA} used a variational auto-encoder for data augmentation.
SC-GPT-NLU \cite{peng2020data} takes the advantage of a massive pre-trained transformer language model (GPT-2 \cite{radford2019language}) in order to augment the variability of the generated utterances.
Table \ref{tab:data_augmentation} reports SC-GPT-NLU \textit{Paired-Data-Only} and \textit{Rich-in-Utterance}.
\textit{Rich-in-Ontology} method is discarded as it uses all the available dataset ontology.

Table \ref{tab:data_augmentation} shows that even our baseline exceeds the performance of SC-GPT-NLU on most SNIPS settings. 
This is due to the optimization of batch size on the development, which must be small $\{4,8\}$, especially for the very low data regimes.
VirAAL random improves the baseline for all metrics with a large gain in Slot F1 for ATIS Small.
VAT joint-random shows better intent accuracy and slot F1 score on all datasets versus our baseline and shows improvements in other methods.
Additionally, experiments show the robustness of VAT across different batch sizes, especially for the slot filling (Std Dev$\approx$9 for ce, Std Dev$\approx$1 for VAT on the validation sets).
As section \ref{results} shows, VirAAL joint-entropy improves intent accuracy over VAT joint-random at the expense of some losses in slot f1.
However, VirAAL individual-entropy gives more consistent improvements over VAT joint-random matching or exceeding in most cases other methods even though  SC-GPT-NLU uses GPT-2, which gives access to a larger and more diverse data.

\section{Conclusion}
\label{conclusion}
This work first demonstrates that Virtual Adversarial Training is a consistent method for training NLU models: intent detection, slot filling and joint training.
This is a very effective way to reduce the amount of annotations to obtain accurate NLU components.
Experiments show that computing a common adversarial noise from multiple loss functions allows to effectively regularize the model even if it was not originally designed for this purpose.
The proposed Virtual Adversarial Active Learning framework, VirAAL, shows even better improvement using entropy-based Active Learning combined with VAT.
This is an inexpensive method in terms of computation for efficiently querying samples to annotate, thanks to the smoothness of the posterior distribution.
Additionally, VirAAL leads to further improvements especially for both intent detection and slot filling and can reduce the labeling effort by up to 80\%.
VirAAL demonstrates a capacity to deal with low data regimes.
When compared with other data augmentation methods, VirAAL outperforms them in most cases, even yielding improvements over augmentation using a large transformer pre-trained language model.
Finally, VirAAL results on joint training suggest that querying samples from a common latent space could improve entropy-based Active Learning rather than combining entropies. 
This could be a future research direction.
\bibliographystyle{IEEEtran}
\bibliography{viraal_bib}

\end{document}